\documentclass[10pt,twocolumn,letterpaper]{article}

\usepackage{wacv}
\usepackage{times}
\usepackage{epsfig}
\usepackage{graphicx}
\usepackage{amsmath}
\usepackage{amssymb}

\usepackage{comment}
\usepackage{multirow, booktabs}
\usepackage{makecell} 
\usepackage{colortbl}
\usepackage[table, dvipsnames]{xcolor}

\usepackage{algorithm2e}
\usepackage{bm}  

\usepackage{caption}
\usepackage{graphicx}
\usepackage{adjustbox}
\usepackage{etoolbox}
\usepackage{soul}
\usepackage{siunitx}
\usepackage{diagbox}
\renewrobustcmd{\bfseries}{\fontseries{b}\selectfont}
\renewrobustcmd{\boldmath}{}
\newrobustcmd{\B}{\bfseries}
\newrobustcmd{\IL}{\underline}

\newcommand{\mb}{\makebox}
\makeatletter
\DeclareRobustCommand\onedot{\futurelet\@let@token\@onedot}
\def\@onedot{\ifx\@let@token.\else.\null\fi\xspace}

\def\eg{\emph{e.g}\onedot} 
\def\ie{\emph{i.e}\onedot} 
 
\def\etc{\emph{etc}\onedot} \def\vs{\emph{vs}\onedot}
 
\def\etal{\emph{et al}\onedot}
\makeatother

\newcommand{\Fig}[1]{Fig. \ref{fig:#1}}
\newcommand{\Eq}[1]{Eq. \ref{eq:#1}}
\newcommand{\Sect}[1]{Sect. \ref{sec:#1}}

\newcommand{\Tab}[1]{Table \ref{tab:#1}}

\usepackage{eucal}

\newcommand{\depthmap}{d}

\newcommand{\mde}{\Psi}

\newcommand{\maxestidepth}{d^{\mbox{{\scriptsize max}}}}
\newcommand{\maxheight}{h^{\mbox{{\scriptsize max}}}}
\newcommand{\minrowdm}{r^{\mbox{{\scriptsize min}}}}

\newcommand{\rgbimage}{I}
\newcommand{\depth}{\depthmap}

\newcommand{\pseudolidardata}{\varrho}
\newcommand{\pl}{\pseudolidardata}
\newcommand{\pseudolidarsampled}{\hat{\varrho}}
\newcommand{\pls}{\pseudolidarsampled}

\newcommand{\posx}{x}
\newcommand{\posy}{y}
\newcommand{\posz}{z}

\newcommand{\numofbeam}{N_b}

\newcommand{\pixelu}{u}
\newcommand{\pixelv}{v}
\newcommand{\princiu}{C_\pixelu}
\newcommand{\princiv}{C_\pixelv}
\newcommand{\focal}{F}
\newcommand{\focalu}{\focal_\pixelu}
\newcommand{\focalv}{\focal_\pixelv}
\newcommand{\objdet}{\Xi}
\newcommand{\intrinsic}{\mathcal{K}}
\newcommand{\lidarsensor}{\mathcal{L}}

\newcommand{\baseexpcolor}{\rowcolor{Gray!40}}

\newcommand{\rowspace}{\\[-1.0em]}  

\newcommand{\angular}{\theta_{\horizontalfov}}
\newcommand{\verticalfov}{\mathcal{V}}
\newcommand{\horizontalfov}{\mathcal{H}}

\def\wacvPaperID{484} 
\wacvfinalcopy 
\ifwacvfinal
\def\assignedStartPage{1} 
\fi

\ifwacvfinal
\usepackage[breaklinks=true,bookmarks=false]{hyperref}
\else
\usepackage[pagebackref=true,breaklinks=true,colorlinks,bookmarks=false]{hyperref}
\fi

\ifwacvfinal
\else
\fi

\begin{document}

\title{\textbf{On the Metrics for Evaluating Monocular Depth Estimation}}

\author{Akhil Gurram\thanks{The majority of the work was done while working at Huawei Munich Research Center}\\
Huawei Munich Research Center, Germany \\
Univ. Aut\`onoma de Barcelona (UAB), Spain\\
{\tt\small akhilgurram.ai@gmail.com}
\and
Antonio M. L\'opez\\
Computer Vision Center (CVC), \\
Univ. Aut\`onoma de Barcelona (UAB), Spain\\
{\tt\small antonio@cvc.uab.es}
}

\maketitle

\begin{abstract}
Monocular Depth Estimation (MDE) is performed to produce 3D information that can be used in downstream tasks such as those related to on-board perception for Autonomous Vehicles (AVs) or driver assistance. Therefore, a relevant arising question is whether the standard metrics for MDE assessment are a good indicator of the accuracy of future MDE-based driving-related perception tasks. We address this question in this paper. In particular, we take the task of 3D object detection on point clouds as proxy of on-board perception. We train and test state-of-the-art 3D object detectors using 3D point clouds coming from MDE models. We confront the ranking of object detection results with the ranking given by the depth estimation metrics of the MDE models. We conclude that, indeed, MDE evaluation metrics give rise to a ranking of methods which reflects relatively well the 3D object detection results we may expect. Among the different metrics, the absolute relative (abs-rel) error seems to be the best for that purpose.
\end{abstract}

\section{Introduction}
\label{sec:introduction}

Monocular depth estimation (MDE) is addressed from different settings determined by the data available at training time, {\eg}, LiDAR \cite{Liu:2016} and virtual-world  \cite{Zheng:2018T2Net} supervision, stereo \cite{Garg:2016} and structure-from-motion (SfM) \cite{Zhou:2017} self-supervision, and combinations of those \cite{Godard:2019MonoDepth2}. MDE results are compared by using \emph{de facto} standard metrics ({\eg}, abs-rel, rms, {\etc}) established by Eigen {\etal} \cite{Eigen:2014}. Reviewing literature results, we can observe that, in terms of such MDE metrics, the difference among different proposals is not too large even the way of training the model is quite different. For instance, we can see it in Table \ref{tab:mde_in_kitti_3dob_validation_high_resolution}. Focusing on the abs-rel metric by now, we can see that for MonoDEVS-SfM abs-rel=0.090, while for MonoDELS-SfM abs-rel=0.077. The former is based on virtual-world supervision and SfM self-supervision \cite{Gurram:2021}, while the latter uses LiDAR supervision instead of the virtual-world one \cite{Gurram:2022-arxiv}. Thus, in terms of physical sensors, the former requires a monocular systems at training time, while the latter requires a relatively dense LiDAR and a camera both calibrated and synchronized. Thus, in our opinion, a reasonable question is whether the difference of $\sim0.13$ points justify the use of a LiDAR-based setting. In fact, when performing MDE on-board an autonomous or assisted vehicle, obtaining depth estimation maps is just an intermediate step of a perception stack. Thus, one may wonder if those differences on depth estimation will be consolidated once such MDE models are used to support the targeted perception task. The aim of this paper is to do a step forward to answer this question, using 3D object detection as downstream perception task.

More specifically, we use already trained and publicly available MDE models from the state-of-the-art to generate depth maps. These depth maps are then used to generate 3D point clouds, termed as Pseudo-LiDAR \cite{Wang:2019} in analogy with LiDAR point clouds. Pseudo-LiDAR is used for training and testing 3D object detectors. We compare the ranking of MDE models according to their performance estimating depth, and their performance supporting 3D object detection through the generation of Pseudo-LiDAR. We consider eight MDE models as well as three different CNN architectures for 3D object detection (Point R-CNN~\cite{Shi:2019}, Voxel R-CNN~\cite{Deng:2021}, CenterPoint~\cite{Yin:2021}). After analysing our experimental results, based on KITTI benchmark~\cite{Geiger:2013}, we have seen that the abs-rel metric is well aligned with 3D object detection results in terms of ranking the MDE methods. What remains as future work is to investigate if we can predict accuracy improvements in 3D object detection from the improvements observed in the abs-rel metric. Otherwise, we recommend to incorporate 3D object detection as part of the evaluation of MDE models. 

\Sect{rw_3DOD} summarizes the related literature. \Sect{methods} presents the models and methods used in our experimental work, namely, the MDE models, the 3D object detection architectures, and the procedure to generate Pseudo-LiDAR from depth maps. \Sect{experimental_results} describes the quantitative and qualitative results, including the performance of the MDE models for both estimating depth and supporting 3D object detection. This allows to compare the rankings of MDE generated by MDE metrics {\vs} 3D object detection metrics. Finally,~\Sect{conclusions} summarizes the presented work.

\section{Related Work}
\label{sec:rw_3DOD}


We focus on MDE methods and 3D object detectors. 

\subsection{Monocular Depth Estimation (MDE)}
As can be seen in the survey by de Queiroz {\etal} \cite{De:2021}, deep convolutional neural networks (CNNs) have significantly boost MDE performance. We can categorize the different MDE proposal according to the data required at training time. For instance, many proposals assume the availability of camera and LiDAR calibrated data, thus, densified LiDAR depth maps are used as supervision to train CNN-based MDE models \cite{Liu:2016, Cao:2017, Xu:2018, You:2019, Bhat:2021}. When available, pixelwise semantic and LiDAR supervision have been used together to improve the performance of MDE models \cite{Gurram:2018}. To avoid the use of LiDAR data, stereo rigs have been used to provide depth self-supervision \cite{Garg:2016, Godard:2017}, also combined with semantic supervision \cite{Chen:2019}. Other works propose stereo self-supervision to improve upon LiDAR-only supervision \cite{Kuznietsov:2017}, assuming a sparse LiDAR setting. In order to avoid the training-time requirement of having access to either a calibrated camera-LiDAR suite or a stereo rig, depth self-supervision has also been based on structure-from-motion (SfM) principles. In other words, only on-board monocular sequences are required to train the corresponding MDE model \cite{Zhou:2017, Yin:2018GeoNet, Zhao:2020}. Since by using SfM supervision alone we can only estimate relative depth, stereo and SfM self-supervision have also been combined \cite{Godard:2019MonoDepth2}, so still keeping a vision-only setting at training time. Other approaches to provide absolute depth in SfM self-supervision settings have been the use of complementary supervision such as the ego-vehicle speed (available as a readable car signal) \cite{Guizilini:20203D}, or synthetic images with depth and semantic supervision \cite{Gurram:2021}. On the other hand, SfM self-supervision has been also used to improve LiDAR-only supervision \cite{Gurram:2022-arxiv}. Finally, it is also worth to mention that many approaches \cite{Kundu:2018AdaDepth, Zheng:2018T2Net, Zhao:2019GASDA, Pnvr:2020SharinGAN, Gurram:2021} explore the use of synthetic images with automatically generated depth supervision, which implies addressing the synth-to-real domain gap.

\subsection{3D Object Detection}
While there are proposals on pure image-based 3D object detection \cite{Li:2019GS3D}, for our study, we are interested in point cloud based object detection. A priori, the point clouds can come from LiDAR sensors or stereo rigs, however, the former case is the most usual in the literature. In any case, we can categorize point cloud based 3D object detection, according to the manner the point cloud is represented. Essentially, we can find point-based and voxel-based approaches. 

\textbf{Point-based.} These methods work directly on the raw point clouds, so preserving the geometric information of the world. PointNet \cite{Qi:2017pointnet} processes point clouds for 3D object recognition and segmentation, while PointNet++~\cite{Qi:2017} runs PointNet in a hierarchical manner to capture local structures and granular patterns of the point cloud. Then, inspired by the popular 2D object detection methods Faster R-CNN \cite{Ren:2017} and SSD \cite{Liu:2016SSD}, the 3D object detectors Point R-CNN~\cite{Shi:2019} and 3D SSD~\cite{Yang:2020} have been proposed, which use PointNet++ as a backbone feature extractor.   

\textbf{Voxel-based.} These methods transform the raw point clouds into a volumetric representation, {\ie}, a  voxel grid. VoxelNet~\cite{Zhou:2018} was the first end-to-end trainable network to learn the informative features by dividing the point cloud into a voxel grid. The grid is processed using a voxel feature encoding (VFE) network to extract 3D features. These features are fed to a region proposal network (RPN) to obtain object probability scores and regress 3D BBs. In fact, VoxelNet has been used as a backbone network in several 3D object detection architectures. For instance,  
Voxel R-CNN~\cite{Deng:2021} uses VoxelNet as part of a RPN. Then, given the corresponding 3D features and 3D BBs, a voxel grouping is performed by a process called \emph{voxel query} (inspired in Ball query~\cite{Qi:2017}). Finally, a PointNet-inspired process is applied to obtain grid point features, which are fed to fully connected networks to perform the final refinement and classification of the BBs. CenterPoint~\cite{Yin:2021}, inspired by CenterNet~\cite{Zhou:2019}, transforms the point cloud into voxels or pillars, using either VoxelNet or PointPillars~\cite{Lang:2019}, respectively. When using VoxelNet as backbone network, the extracted 3D features are flatten to obtain 2D features. Then, a keypoint detector is applied on these 2D features to find the center of potentially detected objects. After, an anchor-free network produces heatmaps to extract additional object properties such as 3D size and orientation. In addition, a light-weighted point network extracts point features at the center of each side of each of the 3D BBs provided by VoxelNet. These point features and those from the object center point are concatenated to pass through MLPs which provide the classification score and 3D BB of each potential object.
\section{Methods}
\label{sec:methods}

\begin{figure}
    \centering
    \includegraphics[width=\columnwidth]{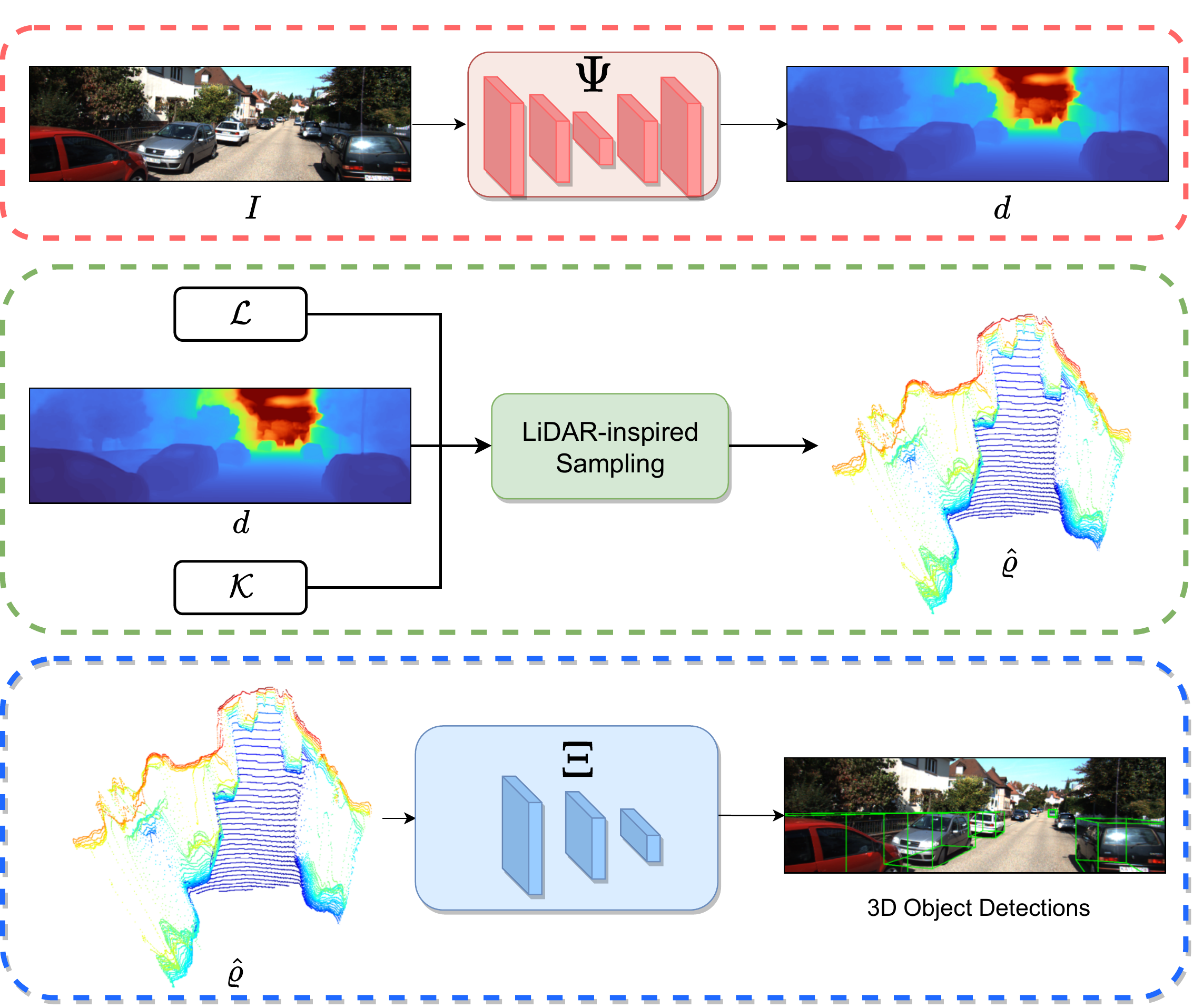}
    \caption{MDE-based 3D object detection in a nutshell. Given an image ($\rgbimage$) we apply a MDE model ($\mde$) to generate its corresponding depth map ($\depth$). With the intrinsic parameters, $\intrinsic$, of the camera capturing the images, as well as a set of parameters, $\lidarsensor$, defining a LiDAR-inspired sampling procedure, the depth map can be converted to a 3D point cloud, $\pls$. Then, $\pls$ is used to train and test a 3D object detector ($\objdet$) originally developed to work with LiDAR point clouds.} 
    \label{fig:pseudo_lidar}
\end{figure}

In order to assess the usefulness of MDE methods for performing 3D object detection, we consider different MDE approaches and 3D object detectors which work on point clouds. While LiDAR already provides such point clouds, for MDE we have to produce the so-called Pseudo-LiDAR point clouds \cite{Wang:2019} by properly sampling the respective depth maps. \Fig{pseudo_lidar}, briefly illustrates the overall idea of performing 3D object detection from monocular images.

\subsection{Pseudo-LiDAR Generation}
\label{sec:produce_pseudo_lidar}
\subsubsection{From a depth map to a 3D point cloud: $\pl$}
In order to generate a Pseudo-LiDAR point cloud, $\pl$, from a depth map, $\depth$, estimated from an image, $\rgbimage$, we need the intrinsic parameters,  $\intrinsic$, of the camera that generated this image. More specifically, we need its optical center $(\princiu, \princiv) $ and focal length\footnote{In practice, calibration software allows to estimate a different focal length parameters per image axis, {\ie}, $\focalu$ and $\focalv$. However, it is expected that $\focalu\approx\focalv$, since this is basically a numerical trick. Thus, for the sake of simplicity, we keep the idea of using a single focal length parameter.} $\focal$, which can be obtained by well-established camera calibration methods \cite{Long:2019}. Therefore, we have $\intrinsic=\{\princiu,\princiv,\focal\}$. Given this information, we can assign a 3D point, $(\posx, \posy, \posz)$, to each pixel, $(\pixelu, \pixelv)$, of the depth map (an so of the input image) as follows:

\begin{equation}   
    \label{eq:posxyz}
        \centering
        \posz\leftarrow{\depth}_{\pixelu, \pixelv} , \, \posy\leftarrow(\posz/\focal)(\pixelv - \princiv) , \, \posx\leftarrow(\posz/\focal)(\pixelu - \princiu)
\end{equation}

Thus, $\pl$ is generated by applying \Eq{posxyz} in all pixels. 

\subsubsection{Sampled Pseudo-LiDAR: $\pls$}
\label{sec:sampled_pseudo_lidar}
As we will see in \Sect{experimental_results}, 
directly working with $\pl$ drives to poor 3D object detection results. We believe that this is because the design of the state-of-the-art 3D object detectors is biased towards the typical 3D pattern distributions present in point clouds captured by actual LiDARs. Therefore, we introduce a LiDAR-inspired sampling procedure which aims at making Pseudo-LiDAR point clouds to be more similar to LiDAR ones. Note that here we are not addressing a domain adaptation problem, since training and testing data will come from the same domain. Instead, we aim at adjusting our generated point clouds to be better suited for training and testing models such as Point R-CNN, Voxel R-CNN, and CenterPoint. 

\begin{figure}
    \centering
    \includegraphics[width=\columnwidth]{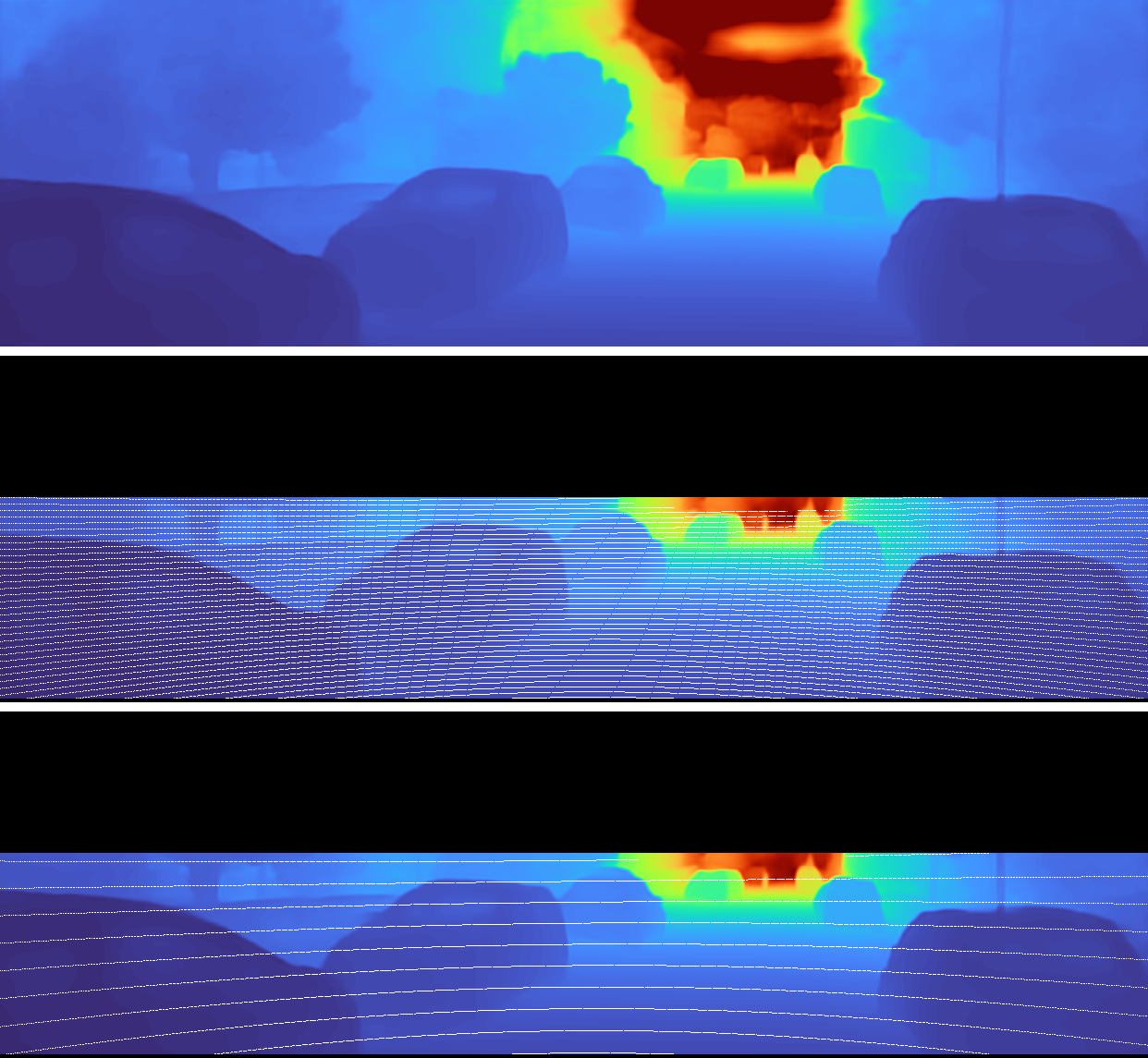}
    \caption{Top: depth map. Mid-Bottom: white pixels would be sampled according to the LiDAR-inspired procedure for $\numofbeam=64$ (mid) and  $\numofbeam=16$ (bottom), where $\numofbeam$ is the number of beams of the simulated LiDAR. }
    \label{fig:sampling}
\end{figure}

Let us introduce the parameters, $\lidarsensor$, required for such LiDAR-inspired sampling. We assume a rotational LiDAR mounted with the rotation axis mainly perpendicular with respect to the road plane. The Velodyne LiDAR HDL 64e used in KITTI dataset, is an example. We term as $\numofbeam$ the number of beams of the LiDAR under consideration. We term as $\verticalfov$ and $\horizontalfov$ the vertical and horizontal field of view (FOV) of the LiDAR, respectively. The vertical angle resolution is $\verticalfov/\numofbeam$, while the horizontal angle resolution, $\angular$, depends on the rotation mechanism. For instance, for the mentioned Velodyne used in KITTI dataset, we have $\numofbeam=64$, $\verticalfov \approx 26.9^{\circ}$, $\horizontalfov=360^{\circ}$, and $\angular\approx 0.08^{\circ}$. Moreover, $\maxestidepth$ and $\maxheight$ denote the maximum depth and height we want to consider above the camera, respectively. Finally, it is common to discard the rows of the depth map above a threshold $\minrowdm$. Therefore, we have $\lidarsensor=\{\numofbeam,\verticalfov,\horizontalfov,\angular,\maxestidepth, \maxheight, \minrowdm\}$. 

\begin{table*}[t!]
    \centering
    \caption{Details about the selected MDE models. } \label{tab:mde_approach_comparison}
    \begin{adjustbox}{width=1\textwidth}
    \small
    \begin{tabular}{|l|c|c|c|c|c|} \Xhline{4\arrayrulewidth}
        \mb[8em]{\textbf{Model}} & \mb[12.5em]{\textbf{Training Data}} & \mb[8.5em]{\textbf{Backbone Encoder}} & \mb[9.5em]{\textbf{Working Resolution (pixels)}} & \mb[8em]{\textbf{Weights (millions)}} \\ 
        \hline \Xhline{4\arrayrulewidth}
        PackNet-SfM              &  Monocular sequences                &   PackNet                             &  $ 1280 \times 384 $                             & 129.88  \\ \hline
        MonoDepth2-St            &  Stereo pairs                       &   ResNet-18                           &  $ 1024 \times 320 $                             &  14.84  \\ \hline
        MonoDepth2-St+SfM        &  Stereo pairs \& Monocular Seq.     &   ResNet-18                           &  $ 1024 \times 320 $                             &  14.84  \\ \hline
        MonoDEVS-SfM             &  Virtual depth \& Monocular Seq.    &   HRNet-w48                           &  $ 1280 \times 384 $                             &  93.34  \\ \hline
        MonoDELS-SfM             &  LiDAR \& Monocular Seq.            &   HRNet-w48                           &  $ 1280 \times 384 $                             &  93.34  \\ \hline
        \mb{MonoDELS-SfM/RN}     &  LiDAR \& Monocular Seq.            &   ResNet-18                           &  $ 1280 \times 384 $                             &  14.84  \\ \hline
        MonoDELS-St                   &  LiDAR \& Stereo pairs              &   HRNet-w48                           &  $ 1280 \times 384 $                             &  93.34  \\ \hline
        AdaBins                  &  LiDAR                              & EfficientNet B5                       & $[1224 - 1242]\times[370 - 376]$                 &  78.25  \\ \hline
        \Xhline{4\arrayrulewidth}
    \end{tabular}
    \end{adjustbox}
\end{table*}

Now we can think of the sampling procedure as follows. We have a \emph{virtual ray} originated in the camera optical center $(\princiu,\princiv)$. This ray samples the image plane (which is at a distance $\focal$ from the principal point), by increments of $\angular$ in its horizontal-component motion, and increments of $\verticalfov/\numofbeam$ in its vertical-component motion. In addition, $\verticalfov, \horizontalfov, \maxestidepth$ and $\maxheight$ set bounds in the 3D space to be considered, while $\minrowdm$ sets a bound in the image space. For the research carried out in this paper, we have set $\verticalfov$ and $\horizontalfov$ so that we consider the full image area, $\maxestidepth=80$m (points with greater depth are not considered), $\maxheight=1$m (points with higher height above the camera are not considered), and $\minrowdm$ is set to discard the top 40\% rows of the depth maps. Note also that the mentioned ray will intersect the image plane in sub-pixel coordinates, however, we take the nearest neighborhood approach to select corresponding pixel coordinates. \Fig{sampling} shows what pixels from a depth map would be sampled to generate the final 3D point cloud following \Eq{posxyz}. 
We term this Pseudo-LiDAR 3D point cloud as $\pls$.

\subsection{MDE models}
\label{sec:mde_models}

In order to generate the Pseudo-LiDAR point clouds, we consider well-established and diverse state-of-the-art methods. Moreover, we prioritize MDE models publicly available, so ready to produce depth maps. Accordingly, based on self-supervision we use MonoDepth2 \cite{Godard:2019MonoDepth2}, PackNet \cite{Guizilini:20203D}, and MonoDEVSNet \cite{Gurram:2021}. Two variants of MonoDepth2 are used, with only stereo self-supervision and with SfM and stereo self-supervision, we will call them MonoDepth2-St and MonoDepth2-St+SfM, respectively. Analogously, since PackNet relies on SfM self-supervision we will term it here as PackNet-SfM. MonoDEVSNet relies on synthetic data supervision and SfM self-supervision, so we will term it here as MonoDEVSNet-SfM.  Based on LiDAR supervision we use AdaBins \cite{Bhat:2021}, and MonoDELSNet \cite{Gurram:2022-arxiv}, which also incorporates SfM self-supervision and, so, we term it here as MonoDELSNet-SfM. We have developed a MDE model by replacing SfM self-supervision with stereo self-supervision on MonoDELS-SfM, we term it as MonoDELS-St.
\Tab{mde_approach_comparison} summarizes the main details related to the selected MDE models.


\subsection{3D object detectors}
\label{sec:train_3d_object_detection}
Regarding 3D object detection, we consider three relatively different approaches which are Point R-CNN \cite{Shi:2019}, Voxel R-CNN \cite{Deng:2021}, and CenterPoint \cite{Yin:2021}. In terms of \Sect{sampled_pseudo_lidar}, any of these 3D object detectors can play the role of $\objdet$. Beyond differences in their respective CNN architectures, an important difference for our study is the fact that they rely on different strategies to represent the 3D point clouds. More specifically, as we have introduced in \Sect{methods}, Point R-CNN assumes a point-based representation, while Voxel R-CNN and CenterPoint rely on a voxel-based representation. Another important point for our research is that these methods have publicly available code for training and testing the respective 3D object detectors.

\section{Experimental Results}
\label{sec:experimental_results}

\subsection{Datasets and evaluation metrics}
\label{sec:datasetsmetrics}
We have downloaded the MDE models reported in \Sect{mde_models}, except for MonoDELSNet-St. The downloaded models were trained on the training set of the Eigen {\etal} \cite{Eigen:2014} split of the KITTI Raw \cite{Geiger:2013} dataset, considering the subset established by Zhou {\etal} \cite{Zhou:2017} when using SfM self-supervision. Accordingly, we have used the same training and testing data for our MonoDELSNet-St. Overall, we can compare all the MDE models since they have been trained on the same data, and will be tested in the same data too.

\begin{figure*}[h]
    \centering
    \includegraphics[width=\textwidth]{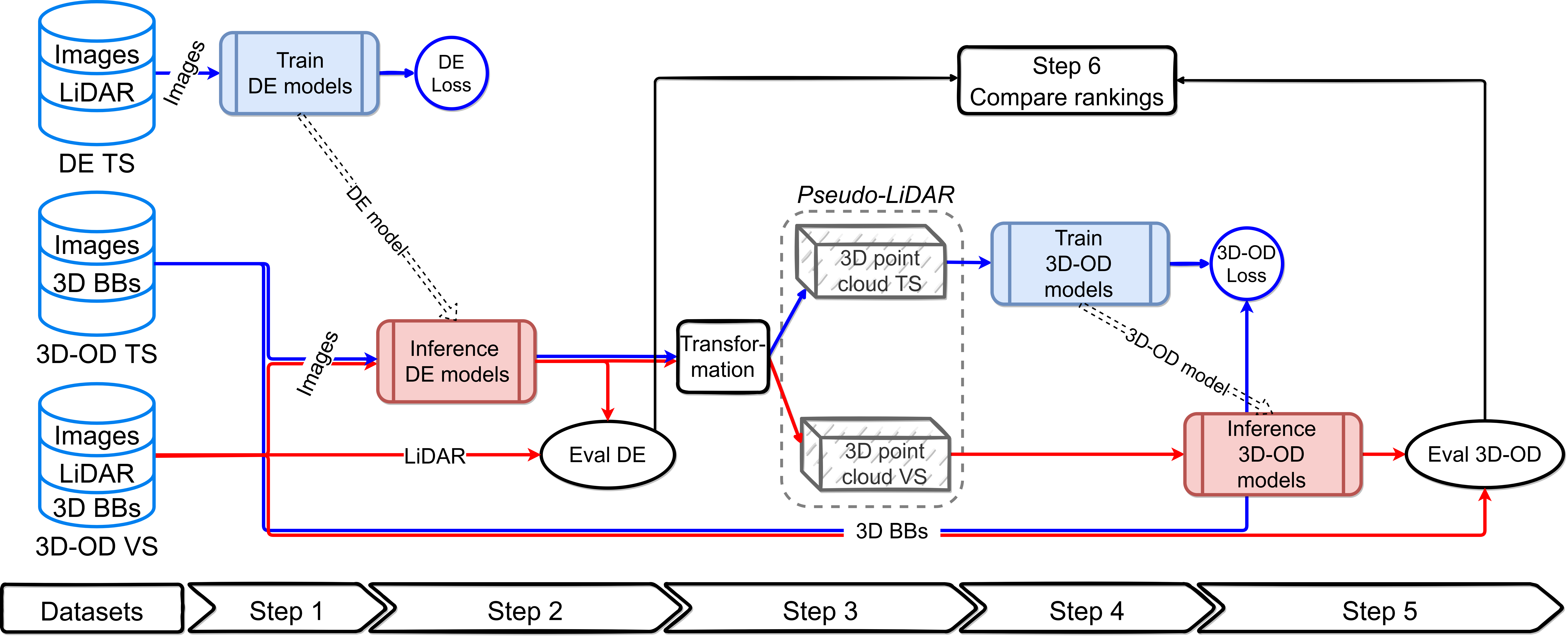}
    \caption{Protocol for experiments. DE stands for depth estimation. TS and VS stands for training and validation set, resp. Here, the DE TS is the training set of the Eigen {\etal} \cite{Eigen:2014} split, the 3D-OD TS and 3D-OD VS are the training and validation sets of the Chen {\etal} \cite{Chen:2015} split. Thus, the datasets are from KITTI benchmark \cite{Geiger:2013}. Eval DE is based on the standard metrics to evaluate depth estimation (abs-rel, rms, ${\delta<\tau}$), while Eval 3D-OD is based on metrics to assess 3D object detection accuracy ($AP_{BEV}$, $AP_{3D}$). Blue paths work at training time, while red paths work at inference/validation time. 
    }
    \label{fig:experiment-protocol}
\end{figure*}

For evaluating 3D object detectors, we consider KITTI object detection benchmark \cite{Geiger:2013}. Moreover, we use the Chen {\etal} \cite{Chen:2015} split, consisting of 3,712 images for training and 3,769 for validation. As we have mentioned before, we use Point R-CNN, Voxel R-CNN, and CenterPoint as 3D object detectors. In order to make the most of the hyper-parameter tuning done by the respective authors, we use their corresponding framework settings. This implies to train Voxel R-CNN only for the class \emph{car}, while Point R-CNN and CenterPoint will also include the classes \emph{pedestrian} and \emph{cyclist}. An important question for our experiments is that this total of 7,481 images come with corresponding 3D point clouds based on a 64-beams Velodyne LiDAR. Camera calibration matrices are also available. Thus, we can obtain the Pseudo-LiDAR depth (\Sect{produce_pseudo_lidar}) by setting $\numofbeam=64$. In the rest of the paper, we term all these data as KITTI-3D-OD.

In this study, we focus on three \emph{de facto} standard and complementary metrics for evaluating depth estimation \cite{Eigen:2014}, defined as follows:  $\textbf{abs-rel} \leftarrow  (1/n)\sum_{ij} |d_{gt} -  d_p |/d_{gt}$, $\textbf{rms} \leftarrow  \sqrt{(1/n)\sum_{ij} (d_{gt} - d_p)^2}$, and, for a threshold $\tau$, $\delta<\tau \leftarrow \text{\% pixels with} \max_{}\left(|d_{p}|/d_{gt}, |d_{gt}|/d_{p} \right) <  \tau,$
where $n$ is the total number of valid pixels ({\ie}, containing depth ground truth) in the full set of testing images, $d_{gt}$ and $d_p$ are the ground truth and predicted depths at a given pixel of any testing image, respectively. The absolute relative error (abs-rel) and the accuracy with threshold $\tau$ (${\delta<\tau}$) are percentage measurements, while the root mean square error (rms) is reported in meters. 

For 3D object detection we use KITTI metrics \cite{Geiger:2013}. More specifically, we report Average Precision (AP) for 3D and bird-eye-view (BEV), {\ie}, AP$_{3D}$ and AP$_{BEV}$. Specific results for the detection-difficulty categories used with KITTI metrics are also reported, {\ie}, for \emph{easy}, \emph{moderate}, and \emph{hard}. For computing these APs, we consider an intersection-over-union (IoU) threshold equal to 0.7 (over 1.0) for the easy case, and 0.5 for the moderate and hard cases. Moreover, as we have mentioned above, the \emph{car} class is the only in common for the default configurations of Point R-CNN, Voxel R-CNN, and CenterPoint. We also trained Voxel R-CNN for the multi-class task. However, performance on pedestrians and cyclists is poor. In fact, for Point R-CNN and CenterPoint neither is so good. Thus, we focus the quantitative analysis on car detection, while qualitative results are presented for the classes that each detector considers.

\subsection{Experiment protocol}
The protocol to conduct our experiments is summarized in \Fig{experiment-protocol}  with the following description of the \textbf{Steps}:

\begin{enumerate}
\item \textbf{Training depth estimation.} MDE models are trained on the \ul{training set of the Eigen et al. split} (from KITTI Raw dataset). As mentioned in \Sect{datasetsmetrics}, we use available trained models (MonoDepth2-Sfm, MonoDepth2-SfM+St, MonoDEVSNet-SfM, MonoDELSNet-SfM, MonoDELSNet-SfM/RN, PackNet-SfM, AdaBins), and train our MonoDELSNet-St.

\item \textbf{Testing depth estimation.} MDE models are applied to the \ul{images of the validation set of the Chen et al. split}. Their performance is assessed with the \ul{LiDAR-based ground truth (depth) of the same validation set}. This is done based on abs-rel, rms, and ${\delta<\tau}$. 

\item \textbf{Generating 3D point clouds.} MDE models are applied to the \ul{images of the training and validation set of the Chen et al. split} (from KITTI-3D-OD dataset). This generates the corresponding depth maps, which are converted into Pseudo-LiDAR as explained in \Sect{produce_pseudo_lidar}. Note that KITTI images and LiDAR data are calibrated, so it is right to assume that we can use the same 3D BBs for LiDAR and Pseudo-LiDAR data. Note that, in case of using an on-board vision-based system to fully replace LiDAR, these 3D BBs should be directly annotated in the Pseudo-LiDAR.

\item \textbf{Training 3D object detection.} The \ul{Pseudo-LiDAR} obtained \ul{from the images of the training set of the Chen et al. split} are used to train the 3D objected detection models summarized in \Sect{train_3d_object_detection}. ({\ie}, Point R-CNN, Voxel R-CNN, CenterPoint). 

\item \textbf{Testing 3D object detection.} The 3D object detectors are applied to the \ul{Pseudo-LiDAR} obtained \ul{from the images of the validation set of the Chen et al. split}. Performance is assessed with $AP_{BEV}$, and $AP_{3D}$.

\item \textbf{Depth estimation {\vs} 3D object detection.} Select a metric on MDE and rank the MDE models accordingly. Assess if this ranking matches with the rankings that would result from the metrics on 3D object detection. The better is the matching, the better the MDE metric is for comparing depth models regarding its expected performance when used for 3D object detection.
\end{enumerate}

%

\begin{table}[b]
    \centering
    \caption{MDE results (up to $80$m) for the images of the validation set of Chen {\etal} \cite{Chen:2015} split. \textbf{Best} and \IL{second best} coding is used. Taking \Fig{experiment-protocol} as reference, abs-rel, rms, and $\delta<\tau$ (here $\tau=1.25$) play the role of Eval DE.}
    \label{tab:mde_in_kitti_3dob_validation_high_resolution} 
    \begin{tabular}{lccc}
        \Xhline{4\arrayrulewidth}  \rowspace
        {\textbf{Model}}&{\mb{\textbf{abs-rel}}}&{\textbf{rms}}&{\bm{$\delta<1.25$}}\\
        \Xhline{4\arrayrulewidth} \rowspace
        PackNet-SfM	      &   0.101   &   4.774   &   0.888     \\ \hline \rowspace
        MonoDepth2-St     &   0.096   &   4.719   &   0.883     \\ \hline \rowspace
        MonoDepth2-St+SfM &   0.095   &   4.609   &   0.892     \\ \hline \rowspace
        MonoDEVS-SfM      &   0.090   &   4.107   &   0.903     \\ \hline \rowspace
        MonoDELS-SfM      & \IL{0.077}&   3.837   &   0.911     \\ \hline \rowspace
        MonoDELS-SfM/RN   &   0.079   &   3.904   &   0.909     \\ \hline \rowspace
        MonoDELS-St	          & \B0.073   & \B3.761   & \IL{0.918}  \\ \hline \rowspace
        AdaBins	          &   0.080   & \IL{3.821}& \B0.920     \\
        \Xhline{4\arrayrulewidth}
    \end{tabular}
\end{table}

\begin{table}[]
    \centering
    \scriptsize
    \caption{Results on 3D car detection using Point R-CNN, Voxel R-CNN and CenterPoint. LiDAR PC refers to training with actual raw LiDAR 3D point clouds. Thus, this case can be seen as the upper-bound for the rest using MDE-based Pseudo-LiDAR. \textbf{Best} and \IL{second best} coding refers to each 3D object detector block. Taking \Fig{experiment-protocol} as reference, $AP_{BEV}$ and $AP_{3D}$ play the role of Eval 3D-OD.}
    \label{tab:SOTA_KITTI_3dobject_bev_3d} 
    \begin{tabular}{lccccccc}
        \Xhline{4\arrayrulewidth} \rowspace
                    &\multicolumn{3}{c}{$AP_{BEV}$}&\multicolumn{3}{c}{$AP_{3D}$}\\ 
        \Xhline{2\arrayrulewidth} \rowspace
        {Model}		        &	{easy}	& {\B mod.} &	{hard}	&	{easy}	& {\B mod.} &   {hard} 	\\ 
        \Xhline{4\arrayrulewidth} \rowspace      
        \multicolumn{7}{c}{Point R-CNN} \\ 
        \Xhline{2\arrayrulewidth} \rowspace \baseexpcolor
        LiDAR PC            &   90.63   &   89.55   &   89.35   &   90.62   &   89.51   &   89.28   \\ \Xhline{2\arrayrulewidth} \rowspace
        PackNet-SfM	        &   47.86   &   32.53   &   30.50   &   45.74   &   31.39   &   27.84   \\ \hline \rowspace
        MonoDepth2-St       &   64.34   &   40.88   &   37.21   &   59.96   &   39.36   &   32.41   \\ \hline \rowspace
        MonoDepth2-St+SfM   &   58.81   &   37.68   &   31.63   &   53.14   &   35.37   &   30.10   \\ \hline \rowspace
        MonoDEVS-SfM	    &   66.17   &   47.66   &   45.30   &   64.33   &   45.93   &   40.01   \\ \hline \rowspace
        MonoDELS-SfM        & \B75.58   & \B56.07   & \B48.68   & \B71.15   & \B53.31   & \B45.92   \\ \hline \rowspace
        \mb{MonoDELS-SfM/RN}&   70.50   &   48.97   &   45.41   &   64.89   &   47.15   &   39.87   \\ \hline\rowspace
        MonoDELS-St              & \IL{74.57}& \IL{54.35}& \IL{47.25}& \IL{68.13}& \IL{47.42}& \IL{42.84}\\ \hline \rowspace
        AdaBins	            &   70.86   &   48.31   &   41.12   &   65.66   &   45.98   &   39.56   \\ 
        \Xhline{4\arrayrulewidth} \rowspace
        \multicolumn{7}{c}{Voxel R-CNN} \\ 
        \Xhline{2\arrayrulewidth} \rowspace \baseexpcolor
        LiDAR PC            &   97.33   &   89.71   &   89.35   &   97.29   &   89.70   &   89.33   \\ \Xhline{2\arrayrulewidth} \rowspace
        PackNet-SfM	        &   55.23   &   37.12   &   35.80   &   52.19   &   34.71   &   33.86   \\ \hline \rowspace
        MonoDepth2-St       &   65.21   &   45.74   &   42.64   &   62.21   &   42.54   &   37.47   \\ \hline \rowspace
        MonoDepth2-St+SfM   &   57.98   &   37.72   &   35.09   &   52.87   &   35.40   &   30.54   \\ \hline \rowspace
        MonoDEVS-SfM	    &   65.73   &   46.90   &   45.24   &   63.06   &   44.69   &   42.84   \\ \hline \rowspace
        MonoDELS-SfM     	& \IL{74.91}& \IL{56.07}& \IL{53.17}& \IL{71.51}& \IL{53.11}& \IL{47.06}\\ \hline \rowspace
        \mb{MonoDELS-SfM/RN}&   69.94   &   52.59   &   46.62   &   65.42   &   47.68   &   43.87   \\ \hline \rowspace
        MonoDELS-St              & \B75.90   & \B57.39   & \B54.23   & \B73.97   & \B55.64   & \B48.73   \\ \hline \rowspace
        AdaBins	            &   70.77   &   52.34   &   46.46   &   66.18   &   47.44   &   43.90   \\ 
        \Xhline{4\arrayrulewidth} 
        \multicolumn{7}{c}{CenterPoint} \\ 
        \Xhline{2\arrayrulewidth} \rowspace \baseexpcolor
        LiDAR PC            &   95.25   &   89.88   &   89.30   &   95.17   &   89.85   &   89.21   \\ 	\Xhline{2\arrayrulewidth} \rowspace
        PackNet-SfM	        &   49.89   &   35.10   &   32.10   &   45.44   &   32.33   &   29.10   \\ \hline \rowspace
        MonoDepth2-St       &   62.14   &   43.22   &   37.59   &   56.43   &   39.82   &   34.96   \\ \hline \rowspace
        MonoDepth2-St+SfM   &   51.00   &   35.12   &   31.72   &   47.01   &   31.50   &   28.12   \\ \hline \rowspace
        MonoDEVS-SfM	    &   61.13   &   44.63   &   42.16   &   56.66   &   41.63   &   37.10   \\ \hline \rowspace
        MonoDELS-SfM	    & \IL{69.23}& \IL{52.87}& \IL{46.46}& 	63.60 	& \IL{48.56}& \IL{43.29}\\ \hline \rowspace
        \mb{MonoDELS-SfM/RN}&   66.72   &   49.65   &   44.82   &   62.57   &   45.65   &   41.14   \\ \hline \rowspace
        MonoDELS-St              & \B74.04   & \B56.14   & \B51.96   & \B68.69   & \B53.58   & \B47.27   \\ \hline \rowspace
        AdaBins	            &   67.92   &   47.21   &   43.04   & \IL{63.62}&   44.85   &   38.63   \\ 
        \Xhline{4\arrayrulewidth}
\end{tabular}
\end{table}

\subsection{Testing MDE and 3D object detection}

MDE results are shown in \Tab{mde_in_kitti_3dob_validation_high_resolution}, while \Tab{SOTA_KITTI_3dobject_bev_3d} present 3D car detection results. The images used to generate the depth maps (and so the Pseudo-LiDAR $\pls$) are from the training set of the Chen {\etal} \cite{Chen:2015} split, while the validation is performed on the validation set of this split too. Concerning 3D object detection, we can see that MonoDELSNet-SfM and MonoDELSNet-St are consistently outperforming the rest of MDE models. The results are still significantly far from the upper-bound based on LiDAR point clouds. Considering only the detection results corresponding to the use of actual LiDAR point clouds, voxel-based methods ({\ie}, Voxel R-CNN and CenterPoint) outperform Point R-CNN. In this case, CenterPoint and Point R-CNN can be actually compared since they consider the same three classes (although quantitative results are reported only for car detection here). However, Voxel R-CNN was tuned for car detection only, so they may outperform the others because of this. On the other hand, this is not crucial here since we compare rankings of MDE {\vs} 3D object detection. 

\begin{figure}
    \centering
    \includegraphics[width=\columnwidth, clip=true, trim=0 0 0 0 ]{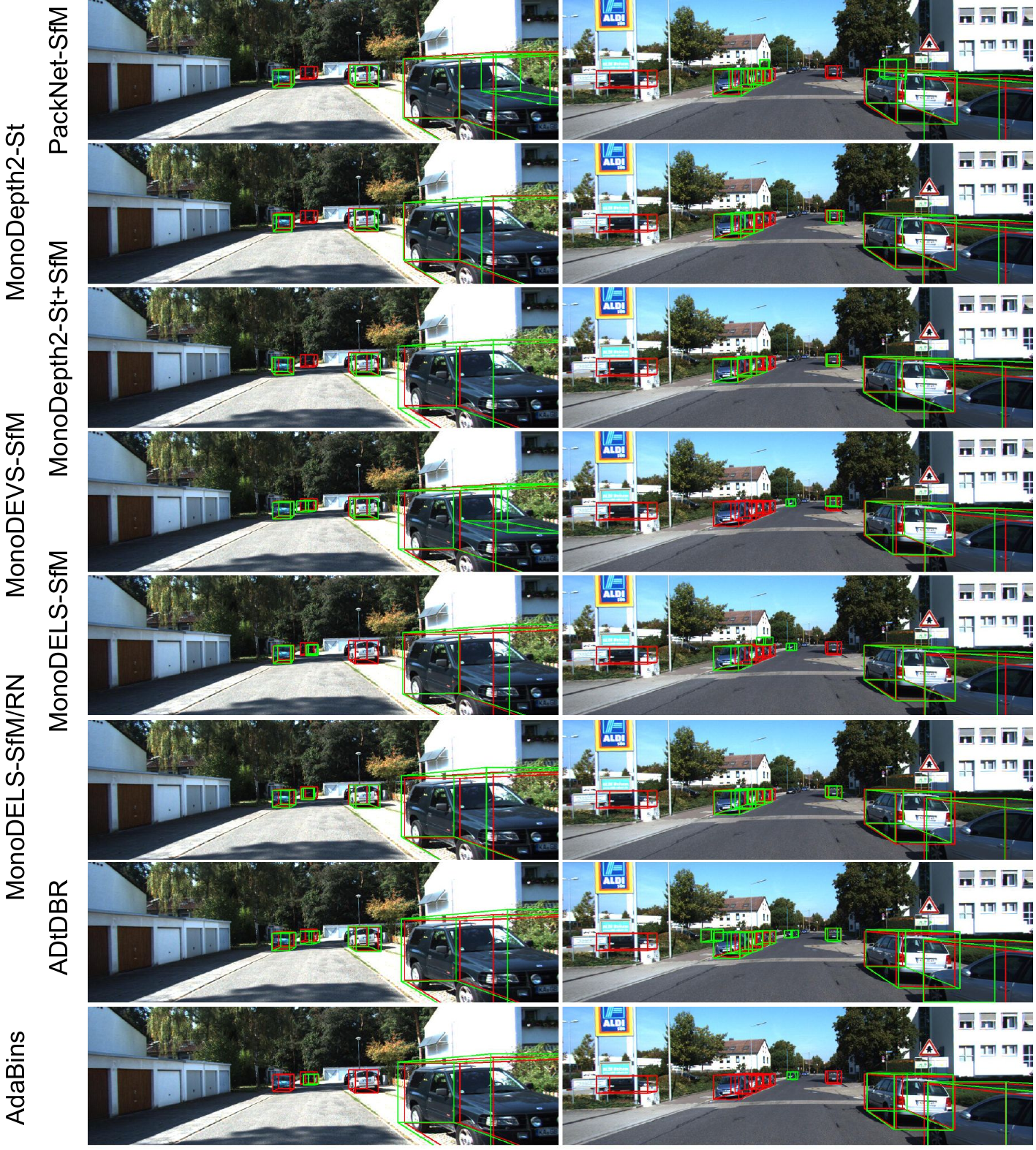}
    \includegraphics[width=\columnwidth, clip=true, trim=0 0 0 0 ]{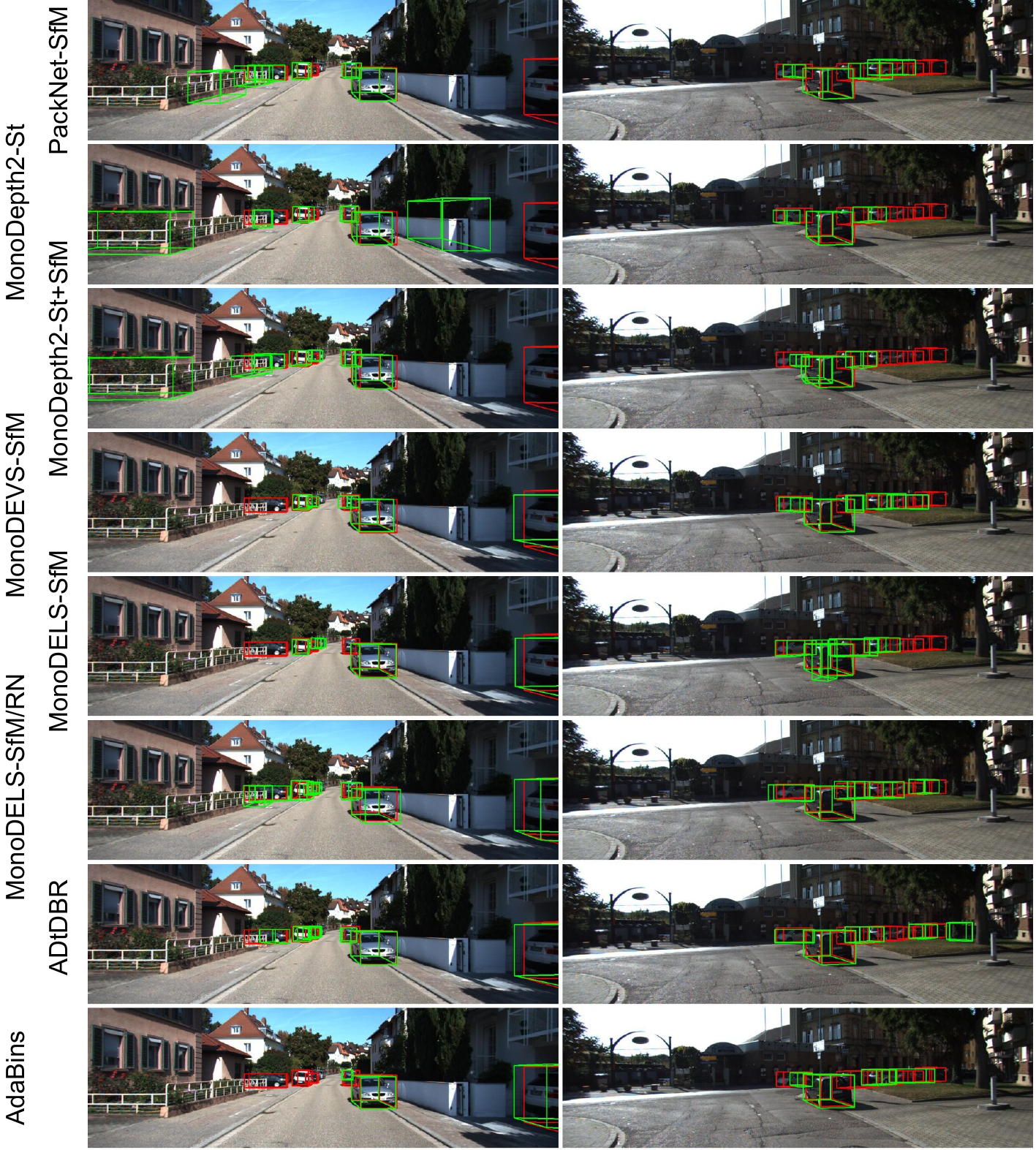}
    \caption{3D object detection for different MDE methods combined with Point R-CNN. Red BBs are {\color{red} ground truth}, green ones are {\color{green} detections}.}
    \label{fig:qual_mde_comparison}
\end{figure}
\begin{figure}
    \centering
    \includegraphics[width=\columnwidth, clip=true, trim=0 390 0 0 ]{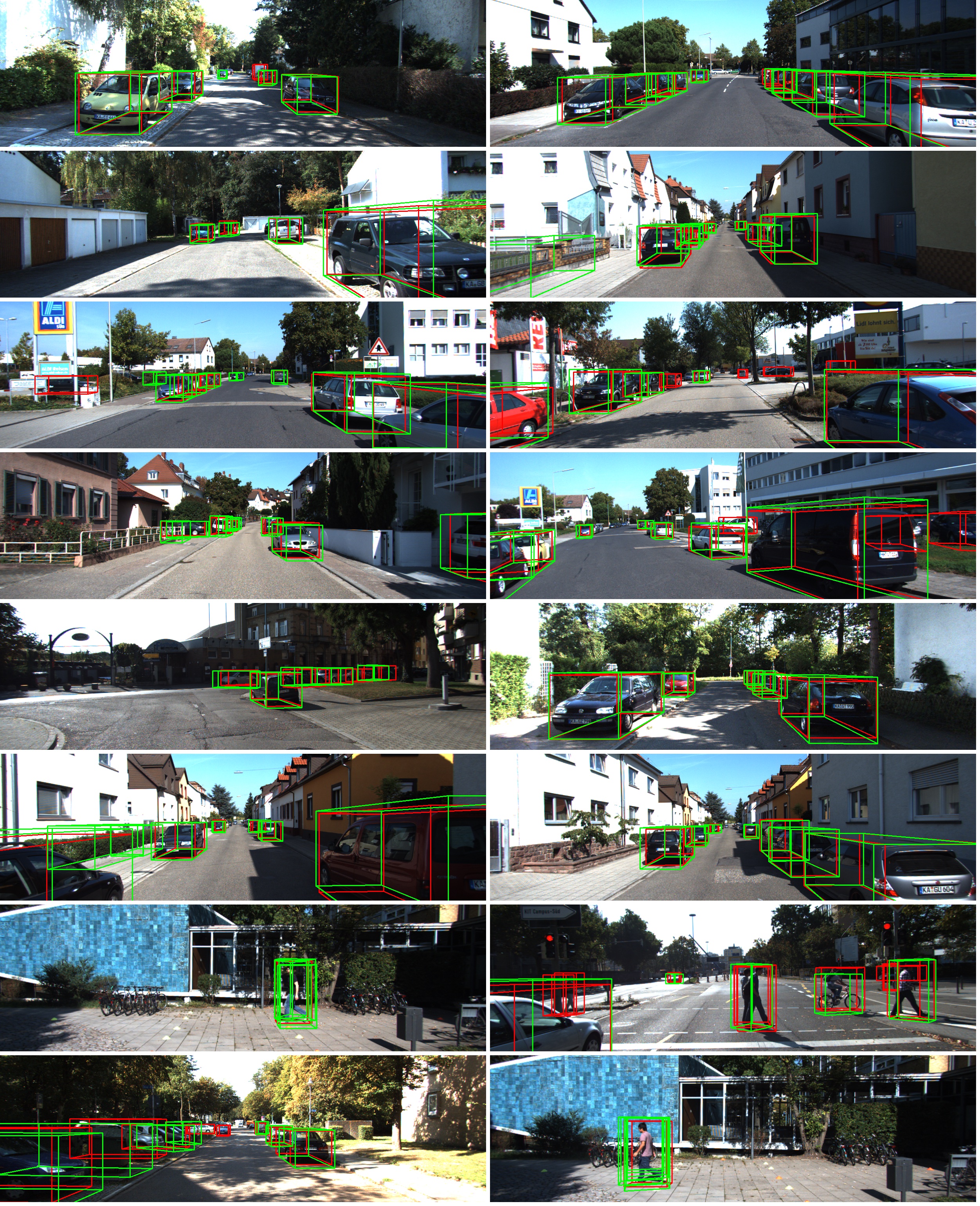}
    \caption{3D object detection based on MonoDELSNet-St combined with Point R-CNN.}
    \label{fig:qual_point_rcnn}
\end{figure}
\begin{figure}
    \centering
    \includegraphics[width=\columnwidth, clip=true, trim=0 390 0 0 ]{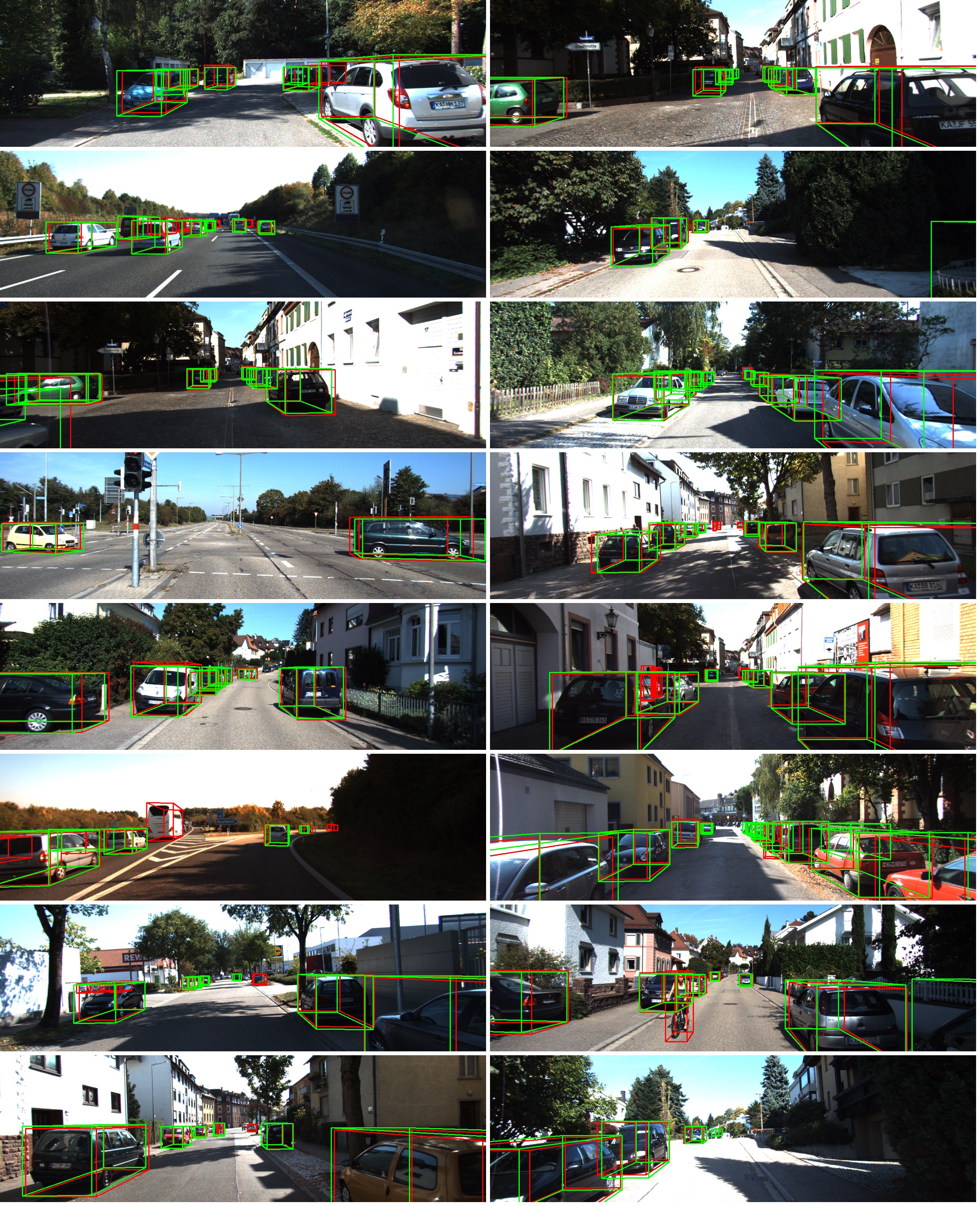}
    \caption{3D object detection based on MonoDELSNet-St combined with Voxel R-CNN.}
    \label{fig:qual_voxel_rcnn}
\end{figure}
\begin{figure}
    \centering
    \includegraphics[width=\columnwidth, clip=true, trim=0 390 0 0 ]{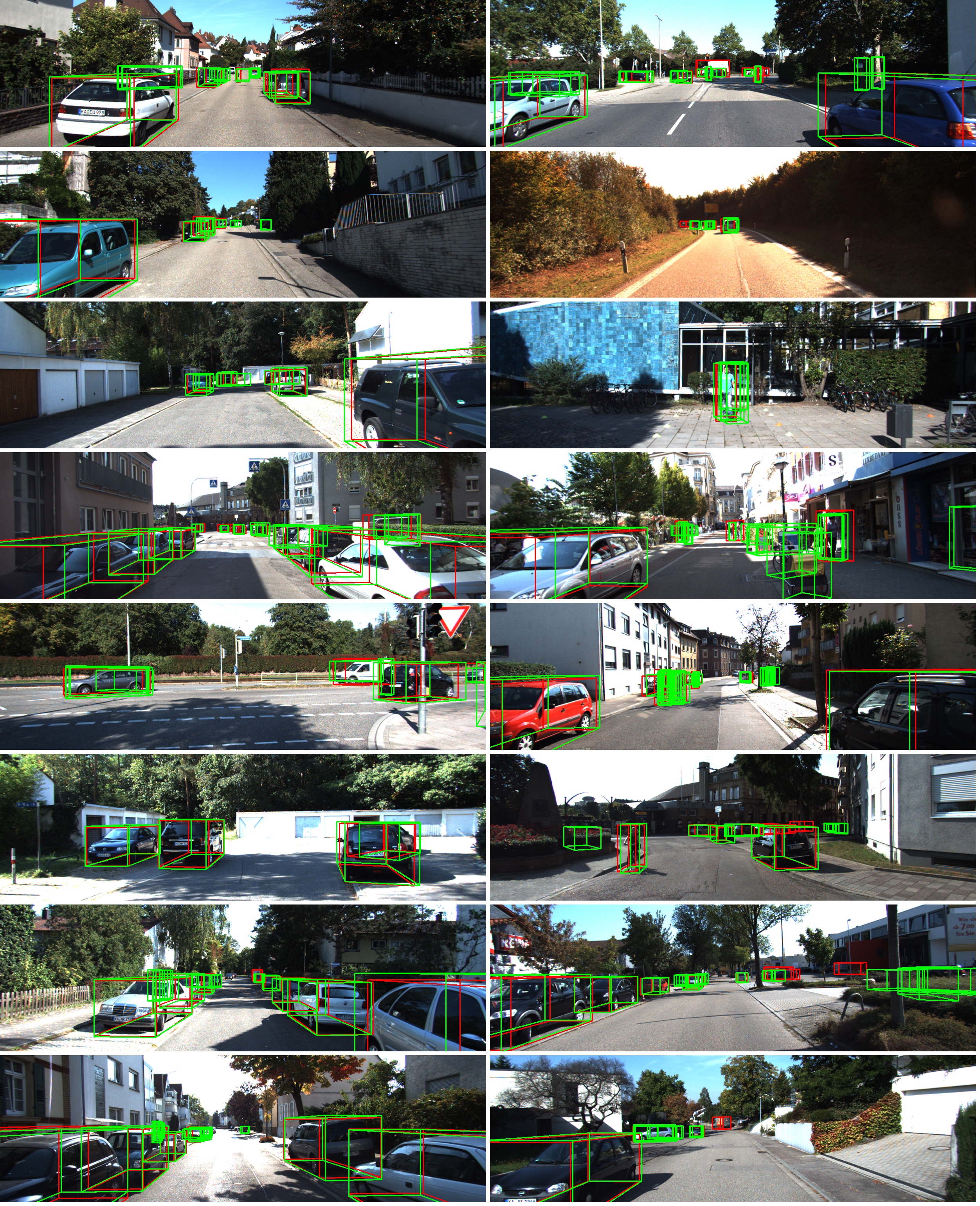}
    \caption{3D object detection based on \mb{MonoDELSNet-St} combined with CenterPoint.}
    \label{fig:qual_centerpoint}
\end{figure}

In addition to the quantitative analysis,  \Fig{qual_mde_comparison} presents qualitative results in terms of 3D object detection for the different MDE methods combined with Point R-CNN. Moreover, Figures \ref{fig:qual_point_rcnn}, \ref{fig:qual_voxel_rcnn}, and \ref{fig:qual_centerpoint}, show results for the MonoDELSNet-St model when using Point R-CNN, Voxel R-CNN, and CenterPoint, respectively. Despite there are errors (false positives and negatives), we think these are very promising results taking into account we are able to provide relatively accurate 3D object BBs (for the true positives) from single images.

\clearpage

\subsection{Depth estimation {\vs} 3D object detection}

\begin{figure}[t!]
    \centering
    \includegraphics[width=\columnwidth]{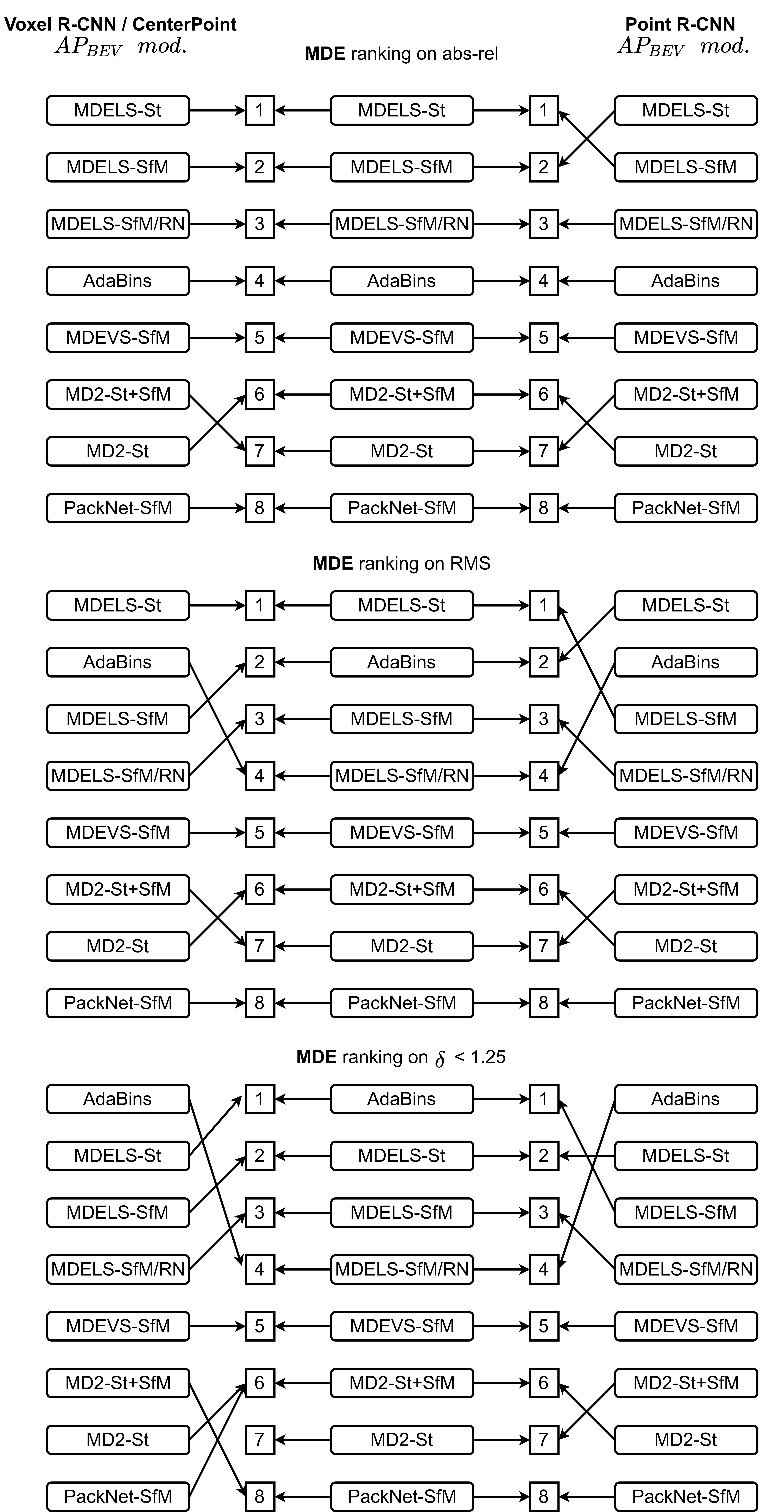}
    \caption{Comparing rankings: abs-rel, rms, $\delta < 1.25$ (MDE) {\vs} $AP_{BEV}-mod$ (3D OD). In the mid column, we have ordered the MDE models from best (top/1) to worse (bottom/8). Then, we have replicated the mid columns as left (voxel-based detectors, which produce the same ranking) and right (Point R-CNN) columns. Afterwards, we have connected the models in left and right columns to its ranking number according to $AP_{BEV}-mod$. Thus, a perfect correspondence between rankings shows as parallel arrows, while the lower the correspondence, the more arrows crossing each other. 
    We abbreviate MonoDepth2 as MD2, MonoDEVSNet as MDEVS, MonoDELSNet as MDELS. }
    \label{fig:ranking_abs_rel}
\end{figure}

Despite examining the results of \Tab{mde_in_kitti_3dob_validation_high_resolution} and \Tab{SOTA_KITTI_3dobject_bev_3d} has a great interest in itself, in this paper, we focus on confronting them. Concerning the 3D object detection metrics, it is common practice to focus on the moderate setting (\textbf{mod.}). Moreover, we see in \Tab{mde_in_kitti_3dob_validation_high_resolution} that $AP_{BEV}$ and $AP_{3D}$ give rise to analogous rankings in terms of 3D object detection under the moderate setting. Thus, for the shake of simplicity, we are going to focus only on $AP_{BEV}$. Accordingly, \Fig{ranking_abs_rel} compares the $AP_{BEV}-mod$ ranking with the abs-rel, rms, and $\delta<\tau$ $(\tau=1.25)$ rankings, for all the considered 3D object detection models. Briefly, the correspondence between MDE and object detection rankings can be visually observed by looking at the arrows in these figures. A perfect correspondence between rankings shows as parallel arrows, while the lower the correspondence, the more arrows crossing each other. Taking this into account, it is clear that the MDE metric abs-rel predicts the 3D object detection results better than the others. The $\delta<\tau$ metric is more messy, and rms metric is performing in between the other two.

\section{Conclusion}
\label{sec:conclusions}
When performing MDE on-board an autonomous or assisted vehicle, obtaining  depth estimation maps is just an intermediate step of a perception stack. For instance, the perception goal may be to perform semantic segmentation or/and object detection with attached depth information to allow for actual vehicle navigation. On the other hand, reviewing the literature of MDE, it is common to see relatively small quantitative differences among the evaluated MDE models. Thus, one may wonder if those differences will be consolidated once such MDE models are used in the targeted perception task. We have addressed this question by using 3D object detection as target perception task. Depth maps based on different MDE models have been converted to Pseudo-LiDAR 3D point clouds, where 3D object detectors can be trained and tested. We have considered eight MDE models, as well as three different CNN architectures for 3D object detection (Point R-CNN, Voxel R-CNN, CenterPoint). Using KITTI benchmark data, we have seen that, indeed, the abs-rel metric commonly used in MDE assessment, is well aligned with 3D object detection results in terms of ranking the MDE methods. What remains as future work is to investigate if we can predict accuracy improvements in 3D object detection (in absolute terms), from the improvements observed in the abs-rel metric. In case this is not possible, we recommend to incorporate 3D object detection as part of the evaluation of MDE models. It is worth to mention that we have also seen that the way depth maps are sampled to produce Pseudo-LiDAR matters because of the dependency that 3D object detectors have on this.

\section*{Acknowledgement}
Antonio acknowledges the financial support received for this research from the Spanish TIN2017-88709-R (MINECO/AEI/FEDER, UE) project. Antonio acknowledges the financial support to his general research activities given by ICREA under the ICREA Academia Program. Antonio acknowledges the support of the Generalitat de Catalunya CERCA Program as well as its ACCIO agency to CVC’s general activities.

We thank Antoni Bigata Casademunt for assistance with sampling Pseudo-LiDAR. We thank Gabriel Villalonga and Onay Urfalioglu for helpful discussions.

\bibliographystyle{ieee_fullname}
\bibliography{main}


\end{document}